\pdfoutput=1

\documentclass[11pt]{article}

\usepackage[preprint]{acl}
\usepackage{etoolbox}
\newtoggle{isSubmission}
\togglefalse{isSubmission}

\usepackage{times}
\usepackage{latexsym}

\usepackage[T1]{fontenc}

\usepackage[utf8]{inputenc}

\usepackage{microtype}

\usepackage{inconsolata}

\usepackage{graphicx}

\usepackage{amsmath,amsfonts,bm}

\def\eqref#1{equation~\ref{#1}}

\def\1{\bm{1}}

\def\rvx{{\mathbf{x}}}
\def\rvy{{\mathbf{y}}}

\DeclareMathAlphabet{\mathsfit}{\encodingdefault}{\sfdefault}{m}{sl}
\SetMathAlphabet{\mathsfit}{bold}{\encodingdefault}{\sfdefault}{bx}{n}

\newcommand{\E}{\mathbb{E}}

\usepackage{url}
\usepackage{multirow}
\usepackage{fancyvrb}
\usepackage{booktabs}
\usepackage{amssymb}
\usepackage{pifont}
\usepackage{hyperref}
\usepackage{stmaryrd}
\usepackage{amsmath}
\usepackage{algorithm}
\usepackage{algpseudocode}
\usepackage[most]{tcolorbox}
\usepackage{listings}
\usepackage{subcaption}
\usepackage{longtable}
\usepackage{colortbl}
\usepackage{pgfplots}
\pgfplotsset{compat=1.17} 

\title{Iterative Length-Regularized Direct Preference Optimization:\\A Case Study on Improving 7B Language Models to GPT-4 Level}

\author{
Jie Liu$^{\ast1,2}$, 
Zhanhui Zhou$^{\ast2}$, Jiaheng Liu$^{2}$, Xingyuan Bu$^{2}$, \\ \textbf{Chao Yang$^{2}$, Han-Sen Zhong$^{\dag2}$, Wanli Ouyang$^{1,2}$}\\
$^{1}$MMLab, CUHK \ $^{2}$Shanghai AI Laboratory
\\
$^\ast$Equal contribution \ \  $^\dag$Corresponding author \\
\texttt{jieliu@link.cuhk.edu.hk}\ \ \ \texttt{asap.zzhou@gmail.com} \\{\tt zhonghansen@pjlab.org.cn}\\
\url{https://huggingface.co/jieliu/Storm-7B}\\
}

\begin{document}
\maketitle
\begin{abstract}
Direct Preference Optimization (DPO), a standard method for aligning language models with human preferences, is traditionally applied to offline preferences. Recent studies show that DPO benefits from iterative training with online preferences labeled by a trained reward model. In this work, we identify a pitfall of vanilla iterative DPO - improved response quality can lead to increased verbosity. To address this, we introduce iterative length-regularized DPO (iLR-DPO) to penalize response length. Our empirical results show that iLR-DPO can enhance a 7B model to perform on par with GPT-4 without increasing verbosity. Specifically, our 7B model achieves a $50.5\%$ length-controlled win rate against \texttt{GPT-4 Preview} on AlpacaEval 2.0, and excels across standard benchmarks including MT-Bench, Arena-Hard and OpenLLM Leaderboard. These results demonstrate the effectiveness of iterative DPO in aligning language models with human feedback.
\end{abstract}

\section{Introduction}

Direct Preference Optimization~\citep{rafailov2024direct} is a standard approach for learning from human feedback~\citep{stiennon2020learning}. 
While DPO typically applies to static offline preferences, recent work~\cite{xu2023some, viethoangtranduong, yuan2024self, xiong2023iterative, xu2024dpo} found that DPO also benefits from iterative online training, where training iterations are interleaved with online preference collection from a reward model.

In this work, we present a case study showing that iterative DPO (iDPO) can enhance a 7B model to GPT-4 level with \textit{careful design}. We make three key contributions: (1) We identify a pitfall of vanilla iDPO -- improved response quality leads to increased verbosity -- a common issue of DPO~\cite{park2024disentangling}, which we find \textit{more critical in multi-iteration online training}.
(2) To address this, we introduce a multi-objective extension of DPO~\cite{zhou2023beyond, park2024disentangling} to penalize response length, termed iterative length-regularized DPO (iLR-DPO).
(3) We empirically show that iLR-DPO outperforms strong baselines in aligning language models. Specifically, iLR-DPO produces a \textit{state-of-the-art} 7B open-source model, achieving a $50.5\%$ length-controlled win rate against \texttt{GPT-4 Preview} on AlpacaEval 2.0~\citep{dubois2024length} and excelling across standard benchmarks including MT-Bench~\citep{zheng2024judging}, Arena-Hard~\citep{arenahard2024} and Open LLM Leaderboard~\citep{open-llm-leaderboard}.
These results highlight iLR-DPO's effectiveness in aligning language models with human values while minimizing alignment tax~\cite{ouyang2022training}.
\iftoggle{isSubmission}{}{Additionally, we have open-sourced our trained model to support future research.}
\begin{figure}[t]
    \centering
    \includegraphics[width=1\linewidth]{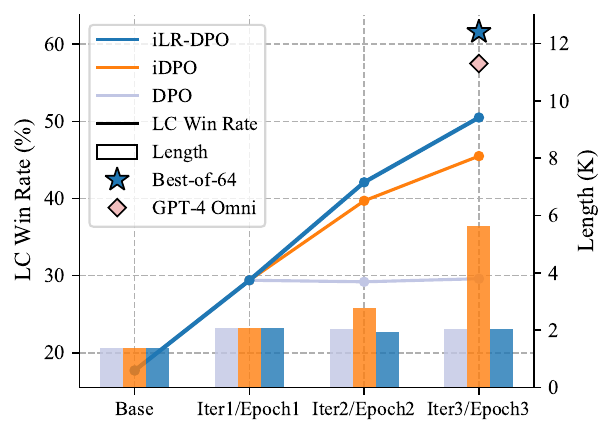}
    \caption{Length-controlled win rates and response lengths on AlpacaEval 2.0. iLR-DPO enhances performance without significantly increasing response length. The trained model achieves a $50.5\%$ length-controlled win rate against \texttt{GPT-4 Preview}, making it the first open-source model to match \texttt{GPT-4 Preview}. 
    }
    \vspace{-2mm}
    \label{fig:teaser}
\end{figure}

\section{Iterative Length-Regularized DPO (iLR-DPO)}\label{sec:method}
In this section, we introduce a simple method to optimize a base language model $\pi_{\text{base}}(\rvy \mid \rvx)$ against a given reward model $r(\rvx, \rvy)$: iterative length-regularized DPO (iLR-DPO). 
The method repeats the following two steps iteratively: 
(1) collect synthetic preferences from the given reward model (Section~\ref{subsec:dataset}) and 
(2) optimize language model on the synthetic preferences with length penalty (Section~\ref{subsec:loss}).

\subsection{Synthetic Preference Collection}\label{subsec:dataset}
For each iteration $i \in \{1, 2, 3, \dots \}$, we first collect synthetic preference feedback from the given reward model $r(\rvx, \rvy)$: 
prompts $\rvx$ are drawn from a prompt set $\mathcal{X}$, 
pair-wise responses $\rvy_1$ and $\rvy_2$ are sampled independently from the latest language model checkpoint $\pi_{\theta_i}(\rvy \mid \rvx)$ for each prompt $\rvx$,
and the preferences between the two responses are annotated by the reward model ($\rvy_1 \succ \rvy_2$ if $r(\rvx, \rvy_1)>r(\rvx, \rvy_2)$). This yields a preference dataset:
\begin{equation}\label{eq:dataset}
    \mathcal{D}_i = \{(\rvx,\rvy_w,\rvy_l)\},
\end{equation}
where $\rvy_w$ are preferred over $\rvy_l$ based on the pre-trained reward model $r(\rvx, \rvy)$.

\subsection{Length-Regularized DPO (LR-DPO)}\label{subsec:loss}
We then optimize the latest language model checkpoint on this synthetic preference dataset using DPO ($\pi_{\theta_i} \rightarrow \pi_{\theta_{i+1}}$). However, language models trained with DPO are prone to generating verbose responses~\cite{park2024disentangling}. Therefore, we use a multi-objective extension to DPO~\cite{zhou2023beyond, park2024disentangling} where we add a length penalty to reduce response verbosity while optimizing for preference. This yields a margin-based cross-entropy loss $\nabla_{\theta_{i+1}}\mathcal{L}_{\text{LR-DPO}}(\pi_{\theta_{i+1}}; \pi_{\theta_i}, \mathcal{D}_i)$:
\begin{align}\label{eq:loss}
    & \nabla_{\theta_{i+1}} \E_{(\rvx, \rvy_w, \rvy_l) \sim \mathcal{D}_i} \left[\log \sigma \left(\beta \mathbf{pm} + 
    \alpha \mathbf{lm}\right) \right]  \\
    & \mathbf{pm} = \log\frac{\pi_{\theta_{i+1}}(\rvy_w\mid\rvx)\pi_{\theta_{i}}(\rvy_l\mid\rvx)}{\pi_{\theta_i}(\rvy_w\mid\rvx)\pi_{\theta_{i+1}}(\rvy_l\mid\rvx)} \nonumber \\
    & \mathbf{lm} = |\rvy_w| - |\rvy_l|, \nonumber
\end{align}
where $\mathbf{pm}$ is the standard preference margin and $\mathbf{lm}$ is the length margin; $|\rvy|$ denotes the length (the number of tokens) of response $\rvy$; $\beta$ and $\alpha$ controls the trade-off between maximizing preferences and minimizing lengths. Training starts from the latest language model checkpoint $\pi_{\theta_{i}}$ and this checkpoint also serves as the frozen reference model in the LR-DPO loss. 

For an intuitive understanding of how Eq.~\ref{eq:loss} controls response length, since $\beta \mathbf{pm} + \alpha \mathbf{lm}$ under different $(\alpha, \beta)$ all represent the same latent preference reward after convergence, positive $\mathbf{lm}$ should therefore lead to a decreased $\mathbf{pm}$ while negative $\mathbf{lm}$ lead to an increased $\mathbf{pm}$.  

\subsection{End-to-End Iterative Training Pipeline}\label{subsection:pipeline}
Denoting the base language model $\pi_{\text{base}}$ as $\pi_{\theta_1}$, we summarize our end-to-end iterative training pipeline as follows:
\begin{equation}
\cdots \rightarrow \underbrace{\pi_{\theta_i} \xrightarrow{\text{Eq.}~\ref{eq:dataset}} \mathcal{D}_i \xrightarrow{\text{Eq.}~\ref{eq:loss}} \pi_{\theta_{i+1}}}_{\text{iteration i}} \rightarrow \cdots.
\end{equation}

\section{Experiments}\label{sec:exp}
In this section, we empirically evaluate iLR-DPO's ability to align language models with human preferences while minimizing alignment tax in various NLP tasks where ground truth answers exist. 

\subsection{Experimental Setup}
\paragraph{Base Model.} We use \texttt{openchat-3.5-0106} \citep{wang2023openchat} as our base model $\pi_{\theta_1}$, which is an open-source language model fine-tuned from \texttt{Mistral-7B-v0.1}~\citep{jiang2023mistral}.

\paragraph{Prompt \& Reward Model.}
We use Nectar~\citep{starling2023}, a preference dataset with diverse chat prompts, high-quality responses, and ranking labels generated by GPT-4. We use these prompts to form our prompt set $\mathcal{X}$ and perform data contamination detection to filter out prompt overlaps with AlpacaEval 2.0~\citep{dubois2024length}. We use \texttt{Starling-RM-34B} as our reward model $r(\rvx, \rvy)$. This reward model is trained on the Nectar dataset.

\paragraph{Evaluation Metrics.}
We assess our models on a standard alignment benchmark, AlpacaEval 2.0~\citep{dubois2024length}, which consists of 805 questions. We report the length-controlled (LC) win rate, a robust metric against model verbosity.  We also evaluate our models on other alignment benchmarks including MT-Bench~\citep{zheng2024judging}, Arena-Hard~\citep{arenahard2024}. We adopt six NLP tasks (including  commonsense reasoning, and math problem solving) from the Open LLM Leaderboard~\citep{open-llm-leaderboard} to measure the "alignment tax", i.e., the performance decrease on traditional NLP tasks with ground-truth answers.

\begin{table*}[h]
\centering
\resizebox{1.0\textwidth}{!}{
\begin{tabular}{lccccc}
\toprule
\multicolumn{1}{c}{\textbf{Model}}             & \textbf{Size}        & \textbf{Open Source} & \textbf{LC Win Rate}     & \textbf{Win Rate}    & \textbf{Avg. Length}    \\ \midrule
iLR-DPO (Ours)          & 7B          & \ding{51}           & \multicolumn{1}{l}{}     & \multicolumn{1}{l}{}   & \multicolumn{1}{l}{}  \\

\qquad Iteration 1       &  &  & $29.4\%$ & $30.5\%$   & $2058$  \\
\qquad Iteration 2       &  &  & $42.1\%$ & $41.7\%$   & $1938$  \\
\qquad Iteration 3       &  &  & $50.5\%$ & $50.3\%$   & $2045$  \\ 
\qquad Iteration 3 + Beam Search $4$       &  &  & $55.1\%$ & $54.6\%$ & $1914$  \\ 
\qquad Iteration 3 + Best-of-$8$      &  & & $58.7\%$  & $59.6\%$ & $2259$ \\ 
\qquad Iteration 3 + Best-of-$64$      &  & & $61.6\%$  & $63.0\%$ & $2340$ \\ \midrule
\textit{Base model} \\
\texttt{openchat-3.5-0106}          & 7B                   & \ding{51}           & $17.7\%$                   & $12.4\%$       & $1376$              \\ \midrule
\textit{Top verified models from the \href{https://tatsu-lab.github.io/alpaca_eval/}{leaderboard}} \\
\texttt{GPT-4 Omni (05/13)} & $\sim$               & \ding{55}                    & $57.5\%$                   & $51.3\%$         &   $1873$       \\
\texttt{GPT-4 Turbo (04/09)}        & $\sim$               & \ding{55}                    & $55.0\%$                   & $46.1\%$         & $1802$         \\
\texttt{GPT-4 Preview (11/06)}      & $\sim$               & \ding{55}                    & $50.0\%$                   & $50.0\%$         & $2049$         \\
\texttt{Llama3-70B-Instruct}              & 70B                  & \ding{51}                    & $34.4\%$                   & $33.2\%$            & $1919$      \\ \bottomrule
\end{tabular}}
\caption{Results on the AlpacaEval 2.0 Leaderboard.}
\vspace{-2mm}
\label{tab:alpacaeval2.0}
\end{table*}

\subsection{Implementation Details}
\paragraph{Training.}
For DPO and iDPO, we perform a grid search for \(\beta\) over \(\{0.01, 0.03, 0.1\}\) for each iteration. For iLR-DPO, we use the same \(\beta\) as iDPO and $\alpha=0.02$. Given that the average response length from most top models on AlpacaEval 2.0 is around $2,000$, we do not apply a length penalty in iteration 1. In subsequent iterations, we add a length penalty to control the response length. More experiment details are in the Appendix~\ref{subsec:training_detail}.
\paragraph{Generation.}
In the first iteration, instead of generating pairwise samples from the base language model ($\pi_{\theta_1} \xrightarrow{\text{Eq.}~\ref{eq:dataset}} \mathcal{D}_1$), we bootstrap from the top two responses from Nectar as $\mathcal{D}_1$. 
The subsequent iterations follow the pipeline in Section~\ref{subsection:pipeline}. 
.

\subsection{Experimental Results}
\paragraph{AlpacaEval 2.0 Leaderboard.}

Table~\ref{tab:alpacaeval2.0} shows that language model's LC win rate improves over iterations without significantly changing the response length, indicating better alignment with human values without length bias. The final trained model (iteration 3) achieves a $50.5\%$ LC win rate, making it the first open-source model to surpass the baseline model \texttt{GPT-4 Preview}.
In addition to regular decoding, we also test beam search and best-of-n sampling on top of our trained model. Beam search over our trained model shows a $5\%$ improvement over regular decoding, Best-of-n sampling with \texttt{Starling-RM-34B} achieves $61.6\%$ LC Win rate and outperforms \texttt{GPT-4 Omni}.

\paragraph{Open LLM Leaderboard.}
Table~\ref{tab:openllm} shows the evaluation results on various tasks from the Huggingface Open LLM Leaderboard. We observe no significant degradation in these traditional NLP tasks with ground-truth answers. Our alignment method improves truthfulness, shown by higher TruthfulQA scores, but reduces performance on math tasks like GSM8K. For other tasks, performance changes are minor.

\begin{table*}[t]
\centering
\resizebox{1.0\textwidth}{!}{
\begin{tabular}{lccccccc}
\toprule
\multicolumn{1}{c}{\textbf{Model}} &
  \multicolumn{1}{c}{\textbf{Avg.}} &
  \multicolumn{1}{c}{\textbf{ARC}} &
  \multicolumn{1}{c}{\textbf{HellaSwag}} &
  \multicolumn{1}{c}{\textbf{MMLU}} &
  \multicolumn{1}{c}{\textbf{TruthfulQA}} &
  \multicolumn{1}{c}{\textbf{Winogrande}} &
  \multicolumn{1}{c}{\textbf{GSM8K}}\\ \midrule
iLR-DPO (Ours)  & & & & & & & \\
\qquad Iteration 1 &$69.89$ &$66.72$ &$80.37$ &$63.04$ &$55.06$ &$80.43$ &$73.69$ \\
\qquad Iteration 2 &$69.02$ &$69.03$ &$78.86$ &$61.37$ &$57.89$ &$80.58$ &$66.41$ \\ 
\qquad Iteration 3 &$68.71$ &$69.11$ &$78.29$ &$61.47$ &$57.57$ &$80.03$ &$65.81$ \\ \midrule
\textit{Base model} \\
\texttt{openchat-3.5-0106} & $69.65$ & $66.30$ & $82.82$ & $63.59$ & $52.52$ & $80.66$ & $72.02$ \\
\bottomrule
\end{tabular}}
\caption{Results on the Open LLM Leaderboard.}
\vspace{-2mm}
\label{tab:openllm}
\end{table*}

\paragraph{Other Instruction-Following Leaderboards.} 
We also evaluate iLR-DPO on MT-Bench and Arena-Hard. MT-Bench has 80 questions across 8 categories, while Arena-Hard includes 500 challenging user queries. Following ~\citet{meng2024simpo}, we use \texttt{GPT-4 Preview} as the judge model in MT-Bench for more accurate answers and judgments than \texttt{GPT-4}. Table~\ref{tab:three_instruction_following} shows that iLR-DPO consistently outperforms iDPO in these benchmarks.

\subsection{Ablation Studies}

\paragraph{Length Penalty.}
Figure~\ref{fig:teaser} shows that for iDPO (without length penalty), both win rate and average response length increase rapidly over iterations. By iteration 3, the length-controlled win rate is $12\%$, far below the raw win rate, and the average length of responses ($5.6$k) is about three times that of GPT-4 ($2$k). 
Overly verbose responses are undesirable as they contain meaningless repetition and overly complex reasoning, consuming unnecessary computational resources. Examples of such responses are in Appendix~\ref{subsec:examples}. In contrast, iLR-DPO can align LLMs more closely with human values without significantly increasing response length.

\begin{table}[]
\centering
\resizebox{0.48\textwidth}{!}{
\begin{tabular}{@{}ccccc@{}}
\toprule
\multirow{2}{*}{\textbf{Method}} & \multicolumn{2}{c}{\textbf{AlpacaEval 2.0}} & \textbf{Area-Hard} & \textbf{MT-Bench}    \\ \cmidrule(l){2-5} 
                        & \textbf{LC (\%)}         & \textcolor{gray}{\textbf{WR (\%)}}        & \textbf{WR (\%)}   & \textbf{GPT-4 Preview} \\ \midrule
Base                     & $17.7$            & \textcolor{gray}{$12.4$}           & $13.0$        & $\underline{6.59}$        \\ \midrule
DPO                     & $29.6$            & \textcolor{gray}{$30.4$}           & \textbf{22.1}      & $6.50$         \\
iDPO                    & $\underline{45.5}$            & \textcolor{gray}{\textbf{57.6}}           & $18.1$      & $6.41$        \\
iLR-DPO                 & \textbf{50.5}            & \textcolor{gray}{$\underline{50.3}$}           & $\underline{20.7}$      & \textbf{7.02}        \\ \bottomrule
\end{tabular}}
\caption{Results on three instruction-following benchmarks. LC and WR denote length-controlled and raw win rate.}
\vspace{-3mm}
\label{tab:three_instruction_following}
\end{table}

\paragraph{Iterative Training vs. Training for More Epochs.}
Figure~\ref{fig:teaser} shows that training DPO on $\mathcal{D}_1$ for more than one epoch is ineffective, as no significant gains occur after the first epoch. In contrast, iteratively generating responses with the latest model, collecting online preferences ($\mathcal{D}_1, \mathcal{D}_2, \mathcal{D}_3$) and training on these preferences prove more effective, despite the higher cost of generating responses.

\paragraph{Achieved Reward.}
We calculate the average reward (\texttt{Starling-RM-34B}) of generated responses for each online iteration of iLR-DPO. The average rewards for $\mathcal{D}_1$, $\mathcal{D}_2$, and $\mathcal{D}_3$ are $-6.57$, $-5.28$, and $-4.31$, suggesting that the language model generates better responses for pair-wise ranking over time, enhancing subsequent training iterations.

\section{Related Work}

\paragraph{Learn from Reward Model.}
Reward models trained on human preferences act as proxies for human preferences.  While some studies propose bypassing explicit reward modeling~\citep{rafailov2024direct}, recent work emphasizes its importance~\citep{fisch2024robust}. Our work supports the latter, assuming access to a reward model from which we collect online preferences as a proxy of human preferences. Specifically, our work, as a case study, demonstrates that using a top ranking reward model \texttt{Starling-RM-34B}~\citep{starling2023} from Reward Bench~\citep{lambert2024rewardbench}, a benchmark for reward models, significantly aligns language models with human values.

\paragraph{Iterative DPO.}
We use ``Iterative DPO'' to describe methods that combine DPO training with online preference collections. These methods can be divided into two categories based on feedback source: (1) languague model feedback, where preferences come from an autoregressive languague model~\citep{guo2024direct, yuan2024self, anil2023palm} and (2) reward model feedback, where preferences are determined by a reward model assuming BT model~\citep{xu2023some, viethoangtranduong, xu2024dpo}. Our method falls into the second category.

\paragraph{Length Regularized Alignment.}
Optimizing for preferences while minimizing verbosity is a multi-objective alignment problem.
MODPO~\citep{zhou2023beyond} introduces a generic margin-based DPO loss to steer language models by multiple objectives. Concurrently with MODPO, \citet{park2024disentangling} analyze the length exploitation in DPO and proposes a more specific (length-)margin-based DPO loss to penalize verbosity. SimPO~\citep{meng2024simpo} uses the average log-likelihood of a response as an implicit reward model. This length-normalized reward formulation prevents length exploitation. All these methods focus on the offline setting.

\section{Conclusion}
We present a case study demonstrating that iterative length-regularized DPO (iLR-DPO) can enhance a 7B model to the GPT-4 level without substantially increasing response length. Our trained 7B model achieves a $50.5\%$ length-controlled win rate on AlpacaEval 2.0, comparable to GPT-4.
\iftoggle{isSubmission}{}{Additionally, we have open-sourced our trained model to support future research.}

\iftoggle{isSubmission}{\section{Limitations}}{\paragraph{Limitations.}}
Our work has several limitations:
(1) We focus on aligning with human preferences but only use GPT-4 as a proxy for human judgment to evaluate language models. 
(2) We reduce verbosity with a length penalty, though verbosity and length are not necessarily correlated. Future work could train a specific reward model to directly penalize verbosity, replacing the length margin with a verbosity margin (Eq.~\ref{eq:loss}), following the standard \href{https://github.com/ZHZisZZ/modpo}{MODPO pipeline}~\cite{zhou2023beyond}.

\iftoggle{isSubmission}{\paragraph{Ethics Statement.}
LLMs might produce offensive or harmful content.}{}

\clearpage
\newpage
\bibliography{all}

\newpage
\appendix
\onecolumn
\clearpage
\noindent

\section{Appendix}
\subsection{Models}\label{subsec:model_link}
The table below lists links to all the language models used in this study.

\begin{table*}[h]
\centering
\small
\begin{tabular}{@{}rl@{}}
\toprule
Models & Model Link \\ \midrule
\texttt{openchat-3.5-0106}  &  \url{https://huggingface.co/openchat/openchat-3.5-0106}               \\
\texttt{Starling-RM-34B}  & \url{https://huggingface.co/Nexusflow/Starling-RM-34B} \\
\texttt{Mistral-7B-v0.1}       &   \url{https://huggingface.co/mistralai/Mistral-7B-v0.1}              \\ 
\texttt{GPT-4 Preview}  &  \texttt{gpt-4-1106-preview} \\
\bottomrule
\end{tabular}
\end{table*}

\subsection{Experiment Details}\label{subsec:training_detail}
We train all models on 8 A100 GPUs with a cosine learning rate scheduler, a learning rate of $5e-6$, an Adam~\citep{kingma2014adam} optimizer, and a global batch size of $48$ for one epoch. We also use LoRA (dropout$=0.05, r=128, \alpha_{\text{LoRA}}=1$). We set $T=1$ to generate chosen and rejected responses.

\subsection{Examples}\label{subsec:examples}

\begin{longtable}{|p{14cm}|}
\toprule
\#\#\# {\bf Question:}\par 
Why can I see the moon during the day?\\\par
\#\#\# {\bf Response:} \par
The moon is an object that is visible for a portion of the day for several reasons, which I'll explain in detail here:\\\par

1. \textbf{Phases of the Moon}: The moon's visibility during the day is largely determined by its phase. During a full moon or a nearly full moon, the moon is bright and almost as luminous as the sun, so it's much easier to see even during the day. However, during the new moon phase, the moon is closer to the sun, and it's not visible from Earth because the sun's light outshines it.\\\par

2. \textbf{Illumination and Contrast}: Even during a waxing crescent or waning crescent phase, the moon can still be visible in the daytime. This is due to a phenomenon called "contrast enhancement," where the moon's illuminated portion has a higher contrast against the dark sky, making it visible even though it's relatively dimmer.\\\par

3. \textbf{Atmospheric Scattering}: When the sun is setting or rising, some of its light gets scattered by our atmosphere, creating a dazzling display of colors known as the sun's iridescence. This scattering of light can make the moon's slightly dimmer glow more noticeable against the sky.\\\par

4. \textbf{Angle of Sun and Moon}: The moon is always in the sky during the day, but its visibility depends on the angle between the sun and the moon. If the moon is low on the horizon, it's less likely to be visible because it's near the sun's glare. However, if the moon is higher in the sky, it can be seen even when it's not fully illuminated.\\\par

5. \textbf{Lunar Eclipses}: During a total lunar eclipse, the Earth's shadow blocks the direct sunlight reaching the moon, but the moon can still be visible in the daytime if it's above the horizon and illuminated by indirect sunlight, known as a "ring of fire" or an annular eclipse.\\\par

6. \textbf{Equinoxes and Solstices}: During the equinoxes and solstices, the sun's path across the sky intersects with the moon's orbit, creating a narrow window when the moon is visible for a longer duration during the day.\\\par

In summary, the moon can be seen during the day due to its phases, contrast, atmospheric scattering, angle, and indirect illumination during certain celestial events. With the right conditions, the moon can even be visible at noon, but it will generally be much less visible when the sun is at its highest point in the sky.\\\par

Finally, it's worth noting that the moon's visibility during the day is also influenced by factors like pollution, humidity, and cloud cover. For example, on a clear day in a polluted city, the moon might be more difficult to spot than on a clear day in a remote area with low humidity. But overall, the main reasons why the moon can be seen during the day are its phase, contrast with the sky, and the angle and position of the sun and moon in relation to each other.\\\par

And, to be more specific, you can see the moon during the day especially well at the time of its first and last quarter (crescent moon) phases. This is because during these phases, the moon and the sun are at right angles to each other, which maximizes the contrast and makes the moon's crescent shape more noticeable. Also, the moon will be much higher in the sky during these phases, which makes it easier to see. At the full moon phase however, the moon would appear much closer to the sun and therefore, it would be more difficult to see. It's worth noting that, while the moon is always orbiting Earth, its position relative to the sun never changes by more than 180 degrees, so the maximum distance between the sun and moon is always less than the moon's diameter. This is why it's difficult to see the moon when it's in a new moon phase, but relatively easy to see it when it's in crescent phases.\\\par

Here's a step-by-step breakdown of why the moon can be seen during the day:\\\par

1. \textbf{Brightness and Phase}: The moon's brightness and phase are the main reasons why it can be seen in the daytime. A full moon or a nearly full moon will appear brighter than the sun, as it reflects more sunlight at once. However, during other phases, such as the crescent moon, the moon is significantly dimmer and harder to see.\\\par

2. \textbf{Angle of Sun and Moon}: The moon's visibility during the day is also affected by the angle between the sun and the moon. If the moon is low on the horizon, it's less likely to be visible because it's near the sun's glare. However, if the moon is higher in the sky, it can be seen even when it's not fully illuminated.\\\par

3. \textbf{Sky Brightness}: At night, the sky is much darker, and the moon is the brightest object in the sky. During the day, the sky is much brighter due to the sun, but the moon's light can still be visible if it's in the right position and has enough contrast against the sky.\\\par

4. \textbf{Earth's Atmosphere}: The Earth's atmosphere can also play a role in the moon's visibility. The atmosphere can scatter sunlight, which can make the moon's light more visible. Additionally, the atmosphere can also scatter the moon's light in different directions, which can make the moon appear bigger or smaller depending on the angle at which it's viewed.\\\par

5. \textbf{Other Factors}: There are other factors that can affect the moon's visibility during the day. For example, humidity, pollution, and cloud cover can all reduce the moon's visibility. Additionally, the moon's distance from Earth can also affect its visibility. The moon is about 384,400 kilometers (238,855 miles) away from Earth on average, and this distance can change depending on the moon's orbit. As the moon gets closer to Earth, it can appear larger and brighter in the sky, making it easier to see during the day.\\\par

Lastly, the moon's visibility during the day is also influenced by the observer's location and the time of year. For example, during a solar eclipse, the moon may be visible in the daytime for several minutes. During a lunar eclipse, the moon is not visible in the daytime, but it can be seen during the night as a red ring around the Earth. In summary, the moon's visibility during the day is influenced by its phase, the angle between the sun and the moon, the sky's brightness, the Earth's atmosphere, and other factors such as the observer's location and the time of year.\\\par

While it's true that the moon is less visible during the day compared to at night, it's still possible to see it in the sky during the day under the right conditions.\\\par

\textbf{To clarify the points above and summarize:}\\\par

- The \textbf{phases} of the moon determine its brightness and visibility during the day. A full moon or nearly full moon will be the easiest to see. At new moon, it is essentially invisible.\\\par

- The \textbf{angle} of the sun and moon in the sky also affects visibility. When the moon is low on the horizon and close to the sun, it is nearly impossible to see. If the moon is higher in the sky, it becomes visible.\\\par

- The \textbf{contrast} between the moon's illuminated portion and the dark sky enhances the moon's visibility during the day.\\\par

- \textbf{Atmospheric scattering} can make the moon's dim glow more noticeable against the sky during the day.\\\par

- The \textbf{moon's orbit} and its distance from the Earth can affect its size and brightness, thus influencing its visibility.\\\par

- Finally, the observer's \textbf{location and time of year} can also affect the moon's visibility. For example, during a solar eclipse, the moon can be visible in the daytime for a short period.\\\par

So, in conclusion, the moon's visibility during the day depends on a combination of its phase, angle, contrast, atmospheric scattering, orbit, and an observer's location and time of year. The moon is visible during the day, but its visibility varies significantly depending on these factors. \\
\bottomrule
\caption{Verbose response generated by iDPO (Iteration3).}
\label{fig:appendix_idpo_example1}
\end{longtable}

\begin{longtable}{|p{14cm}|}
\toprule
\#\#\# {\bf Question:}\par 
Why can I see the moon during the day?\\\par
\#\#\# {\bf Response:} \par
There are several factors that allow you to see the Moon during the day, and here's a detailed explanation:\\\par

1. \textbf{Lunar illumination and phases}: The Moon constantly changes its illuminated portion as it orbits the Earth. During full moon, the entire face visible to us is illuminated, so it's highly visible during daytime. Even in the 'last quarter' (waxing gibbous phase), the Moon is often visible during the day.\\\par

2. \textbf{Moon's brightness}: The Moon is one of the brightest objects in the sky, second only to the Sun. On a clear day, with no direct sunlight obscuring it, the Moon is bright enough to be seen even in the sky's blue daylight.\\\par

3. \textbf{Atmospheric scattering}: Light from the Moon gets scattered by our atmosphere, but it's still bright enough to be seen during the day. The scattering makes the Moon appear in a different color and can create a blue halo around it.\\\par

4. \textbf{Angle of the Sun-Earth-Moon system}: The best daytime visibility occurs when the Sun, Earth, and Moon are nearly in a straight line (or close to a new moon phase). In these cases, the Moon is in the opposite direction of the Sun from our perspective on Earth.\\\par

5. \textbf{Terrestrial factors}: Local weather conditions, the position of the observer, and the time of the year can all impact daytime visibility of the Moon. For instance, during winter solstice, the Moon can be seen high in the sky during daytime because the Earth's axial tilt reduces the Sun's apparent altitude.\\\par

In summary, the Moon is visible during the day because of its brightness, consistent illumination, scattering of light by the atmosphere, and a suitable angle relative to the Sun. It's a common sight to see the Moon during the daytime, but the exact visibility can depend on the phase, local conditions, and the time of the year. \\
\bottomrule
\caption{Concise response generated by iLR-DPO (Iteration3).}
\label{fig:appendix_idpo_example2}
\end{longtable}

\begin{longtable}{|p{14cm}|}
\toprule
\#\#\# {\bf Question:} \par 
How would I go about teaching my cat French?\\\par
\#\#\# {\bf Response:} \par
Teaching a cat French might seem like a fun and unique challenge, but it's important to remember that cats have a different cognitive and learning process compared to humans or dogs. Here's a step-by-step guide to help you teach your cat French:\\\par

1. \textbf{Set realistic expectations}: Cats are not naturally inclined to learn human languages, especially not in the same way as dogs or parrots. However, with patience and consistency, you can teach your cat to associate certain words or phrases with specific actions or objects.\\\par

2. \textbf{Start with basic associations}: Begin by associating French words with objects, actions, or sounds that your cat already understands. For example, if your cat knows the word "food," you can start by saying "poulet" (chicken in French) when you give them their meal. Over time, your cat may start to recognize the word "poulet" as a cue for mealtime.\\\par

3. \textbf{Use positive reinforcement}: Cats respond well to positive reinforcement, so reward your cat with treats, praise, or playtime when they show signs of understanding or responding to a French word or phrase.\\\par

4. \textbf{Create a consistent routine}: Establish a daily routine that includes teaching your cat French words or phrases. Consistency is key to helping your cat learn and retain new information.\\\par

5. \textbf{Use multisensory learning}: Cats have a strong sense of smell and hearing, so use these senses to help them learn French. You can use French-language recordings, scents, or visual cues to help your cat associate words with their meanings.\\\par

6. \textbf{Teach one word or phrase at a time}: Start with simple words or phrases and gradually increase the complexity as your cat becomes more comfortable. It's important not to overwhelm your cat with too much information at once.\\\par

7. \textbf{Be patient and persistent}: Teaching a cat French will take time and patience. Be prepared to spend weeks or even months working on this project. It's important to stay consistent and patient, and to celebrate small victories along the way.\\\par

8. \textbf{Use a clicker}: A clicker can be a useful tool for teaching your cat to associate a specific sound with a reward. You can use the clicker to mark the exact moment your cat responds to a French word or phrase, and then immediately give them a treat or praise.\\\par

9. \textbf{Incorporate playtime}: Cats love to play, so use this to your advantage by incorporating French words into playtime. For example, you can use a French-language toy or teach your cat to fetch a French-language book.\\\par

10. \textbf{Consult a feline behaviorist}: If you're having trouble teaching your cat French, or if you're concerned about their progress, consult a feline behaviorist or a veterinarian who specializes in animal learning. They can provide guidance and support to help you achieve your goal.\\\par

It's important to note that while it's possible to teach your cat to associate certain words or phrases with specific actions or objects, it's unlikely that your cat will become fluent in French in the same way a human or a parrot would. However, with patience, consistency, and a creative approach, you can certainly help your cat learn and understand some French words and phrases.\\\par

Here's a more detailed breakdown of the process:\\\par

- \textbf{Start with simple words}: Begin by teaching your cat to associate a few simple French words with specific actions or objects. For example, you can teach them to respond to "bonjour" (hello) by coming to you, or to associate "chat" (cat) with their food bowl.\\\par

- \textbf{Use repetition and reinforcement}: Repeat the words and phrases consistently, and reward your cat with treats or praise when they respond correctly. This will help them learn to associate the words with the actions or objects.\\\par

- \textbf{Gradually increase complexity}: As your cat becomes more comfortable with simple words, you can start teaching them more complex phrases or sentences. You can use flashcards or a language learning app to help them learn new words and phrases.\\\par

- \textbf{Use positive reinforcement}: Cats respond well to positive reinforcement, so use treats, praise, or playtime to reward your cat when they show signs of understanding or responding to a French word or phrase.\\\par

- \textbf{Create a language-rich environment}: Surround your cat with French-language materials, such as books, recordings, or toys, to help them become more familiar with the language.\\\par

- \textbf{Be patient and consistent}: Teaching your cat French will take time and patience. It's important to stay consistent and patient, and to celebrate small victories along the way.\\\par

- \textbf{Monitor your cat's progress}: Keep track of your cat's progress and adjust your teaching methods as needed. If your cat seems to be struggling with a particular word or phrase, try a different approach or simplify the concept.\\\par

- \textbf{Consider using a language learning app}: There are several language learning apps available that can help you teach your cat French. These apps can provide you with a structured curriculum and help you track your cat's progress.\\\par

- \textbf{Have fun}: Teaching your cat French should be a fun and enjoyable experience for both of you. Use positive reinforcement and playtime to make the learning process enjoyable and rewarding for your cat.\\\par

In summary, while teaching your cat French may be a unique and challenging endeavor, it's certainly possible with patience, consistency, and a creative approach. With the right techniques and a lot of love, you can help your cat learn and understand some French words and phrases.\\\par

Finally, it's important to remember that the ultimate goal is not to have a fluent French-speaking cat, but rather to create a strong bond and a positive learning experience for your feline friend. The process of teaching your cat French can be a fun and rewarding way to strengthen your relationship and deepen your understanding of your cat's unique cognitive abilities.\\\par

Here's a practical example of how you might start teaching your cat French:\\\par

1. \textbf{Choose a simple word or phrase}: Start with a word or phrase that is easy for your cat to understand and associate with a specific action or object. For example, you can start with the word "dîner" (dinner in French).\\\par

2. \textbf{Create a consistent routine}: Establish a daily routine where you say "dîner" when it's time for your cat's meal. Repeat this consistently, and reward your cat with treats or praise when they come to you or show signs of understanding.\\\par

3. \textbf{Use positive reinforcement}: When your cat comes to you in response to the word "dîner," give them a treat or praise them enthusiastically. This will help them associate the word with the positive experience of getting a treat or praise.\\\par

4. \textbf{Gradually increase complexity}: Once your cat has learned to associate the word "dîner" with mealtime, you can start teaching them more complex phrases or sentences. For example, you can say "dîner est prêt" (dinner is ready) when you put their food bowl down.\\\par

5. \textbf{Monitor your cat's progress}: Keep track of your cat's progress and adjust your teaching methods as needed. If your cat seems to be struggling with the word "dîner," try simplifying the concept or using a different approach.\\\par

6. \textbf{Be patient and consistent}: Teaching your cat French will take time and patience. It's important to stay consistent and patient, and to celebrate small victories along the way.\\\par

7. \textbf{Have fun}: Make the learning process enjoyable for both you and your cat. Use positive reinforcement and playtime to make the experience fun and rewarding for your feline friend.\\\par

In conclusion, while teaching your cat French may seem like a unique and challenging endeavor, it's certainly possible with patience, consistency, and a creative approach. With the right techniques and a lot of love, you can help your cat learn and understand some French words and phrases. The ultimate goal is not to have a fluent French-speaking cat, but rather to create a strong bond and a positive learning experience for your feline friend.\\\par

Here's a more detailed example of how you might teach your cat to understand the word "dîner": \\
\bottomrule
\caption{The responses generated by iDPO (Iteration 3) are too verbose and long, resulting in truncation by \texttt{max\_tokens}.}
\label{fig:appendix_idpo_example3}
\end{longtable}

\begin{longtable}{|p{14cm}|}
\toprule
\#\#\# {\bf Question:}\par 
How would I go about teaching my cat French?\\\par
\#\#\# {\bf Response:} \par
Cats are naturally adept at learning through positive reinforcement, association, and repetition. Teaching your cat French involves a multi-step process that combines these principles and caters to their unique learning style. Here's a detailed plan to teach your cat French:\\\par

1. \textbf{Establish a routine}: Set aside dedicated time every day for training sessions, ideally 10-15 minutes at a time. Consistency is key.\\\par

2. \textbf{Start with basic associations}: Begin with simple associations like linking a word or sound with a positive experience (e.g., a treat, petting, or playtime). For French, start with common words like "bonjour" (hello), "chat" (cat), and "ma chatte" (my cat).\\\par

3. \textbf{Use positive reinforcement}: Reward your cat with treats, praise, or playtime immediately after they respond correctly to a command or word. Be consistent with the reward and timing.\\\par

4. \textbf{Associate words with actions}: Teach your cat to associate French words with specific actions. For example, "sauter" (jump) can be paired with a jump command. Use hand signals and body language to help reinforce the association.\\\par

5. \textbf{Gradually increase complexity}: Start with simple phrases and gradually introduce more complex sentences. Use a slow, steady pace to avoid overwhelming your cat.\\\par

6. \textbf{Use a clicker}: A clicker can help mark the exact moment your cat responds correctly, reinforcing the association between the word and the action.\\\par

7. \textbf{Record and play French audio}: Expose your cat to native French speakers and recordings to help them associate the language with real-life contexts.\\\par

8. \textbf{Use a language learning app or program}: There are several apps and programs designed for teaching languages to animals, including cats. These resources can provide structured lessons and reinforce learning through games and rewards.\\\par

9. \textbf{Be patient and persistent}: Cats learn at their own pace, and progress may be slower than with dogs. Be patient and consistent in your training, and celebrate small victories along the way.\\\par

10. \textbf{Consider a professional}: If you're serious about teaching your cat French, consider working with a professional animal behaviorist or a certified cat trainer who specializes in language acquisition.\\\par

Remember, the goal is to make learning French an enjoyable and rewarding experience for your cat. Keep sessions fun and engaging, and always prioritize your cat's well-being and comfort. With dedication and patience, your cat can learn to understand and even respond to French commands and phrases. \\
\bottomrule
\caption{Concise response generated by iLR-DPO (Iteration 3).}
\label{fig:appendix_idpo_example3}
\end{longtable}

\end{document}